\pdfoutput=1

\documentclass[11pt]{article}

\usepackage[]{EMNLP2022}

\usepackage{times}
\usepackage{latexsym}

\usepackage[T1]{fontenc}

\usepackage[utf8]{inputenc}

\usepackage{microtype}

\usepackage{inconsolata}

\usepackage{amsmath,amsfonts,bm}

\def\vx{{\bm{x}}}
\def\vy{{\bm{y}}}
\def\vz{{\bm{z}}}

\DeclareMathAlphabet{\mathsfit}{\encodingdefault}{\sfdefault}{m}{sl}
\SetMathAlphabet{\mathsfit}{bold}{\encodingdefault}{\sfdefault}{bx}{n}

\usepackage{color}
\usepackage{xcolor}
\definecolor{ugreen}{rgb}{0,0.5,0}
\definecolor{lgreen}{rgb}{0.9,1,0.8}
\definecolor{lightgray}{gray}{0.85}
\definecolor{myblack}{rgb}{0.15,0.15,0.15}
\definecolor{lyellow}{rgb}{0.54, 0.25, 0.27}
\definecolor{mypurple}{rgb}{0.6, 0.4, 0.8}

\definecolor{darkblue}{rgb}{0.0, 0.0, 0.55}
\definecolor{darkcandyapplered}{rgb}{0.64, 0.0, 0.0}
\definecolor{babyblue}{rgb}{0.54, 0.81, 0.94}
\definecolor{ballblue}{rgb}{0.13, 0.67, 0.8}
\definecolor{bluemunsell}{rgb}{0.0, 0.5, 0.69}
\usepackage{caption}
\usepackage{enumitem}
\usepackage{amsmath}
\usepackage{amsfonts} 
\usepackage{pgfplots}
\usepackage{tikz}
\usepackage{subfig}
\usetikzlibrary{backgrounds,fit} 
\usetikzlibrary{shapes,arrows,shadows}
\usetikzlibrary{patterns}
\usetikzlibrary{shapes.geometric} 
\usetikzlibrary{decorations.pathreplacing} 
\usetikzlibrary{calc}
\usepackage{array,multirow}
\usepackage{array} 
\usepackage{booktabs} 
\usepackage{arydshln} 
\usepackage{diagbox} 
\usepackage{bbm} 
\newcommand{\PreserveBackslash}[1]{\let\temp=\\#1\let\\=\temp} 
\newcolumntype{C}[1]{>{\PreserveBackslash\centering}p{#1}}
\newcolumntype{R}[1]{>{\PreserveBackslash\raggedleft}p{#1}}
\newcolumntype{L}[1]{>{\PreserveBackslash\raggedright}p{#1}}
\newcolumntype{M}[1]{ >{\centering\arraybackslash}m{#1}}
\usepackage{cleveref}
\usepackage{xspace}
\usepackage{algorithm,algorithmicx,algpseudocode}
\usepackage{multicol}
\usepackage{bm}

\pgfplotsset{compat=1.16}

\usepackage{verbatim}
\usepackage{float}

\newlength{\vseg}
\newlength{\hseg}
\newlength{\wnode}
\newlength{\hnode}
\newcommand{\bos}{[\texttt{BOS}]\xspace}
\newcommand{\shortbos}{[\texttt{B}]\xspace}

\newcommand{\eos}{[\texttt{EOS}]\xspace}
\newcommand{\shorteos}{[\texttt{E}]\xspace}
\newcommand{\pad}{[\texttt{PAD}]\xspace}
\newcommand{\shortpad}{[\texttt{P}]\xspace}
\newcommand{\mask}{[\texttt{MASK}]\xspace}
\newcommand{\shortmask}{[\texttt{M}]\xspace}

\newcommand{\spe}[1]{[\texttt{#1}]\xspace}

\newcommand{\myname}{\textsc{RoyalFlush}\xspace}

\title{The RoyalFlush System for the WMT 2022 Efficiency Task}

\author{
    Bo Qin$^{1}$,
    Aixin Jia$^{1}$,
	Qiang Wang$^{2,1}$, \\
	\textbf{Jianning Lu}$^{1}$,
	\textbf{Shuqin Pan}$^{1}$,
	\textbf{Haibo Wang}$^{1}$,
	\textbf{Ming Chen}$^1$\thanks{\xspace\xspace Corresponding author.} \\
	$^{1}$RoyalFlush AI Research Institute, Hangzhou, China\\
	$^{2}$Zhejiang University, Hangzhou, China\\
	{\tt
		\{qinbo, jiaaixin, wangqiang3\}@myhexin.com
	}\\ 
	{\tt
		\{lujianning, panshuqin, wanghaibo3, chenming\}@myhexin.com
	} \\
}

\begin{document}
\maketitle
\begin{abstract}
This paper describes the submission of the \myname neural machine translation system for the WMT 2022 translation efficiency task. 
Unlike the commonly used autoregressive translation system, we adopted a two-stage translation paradigm called Hybrid Regression Translation (HRT) to combine the advantages of autoregressive and non-autoregressive translation.
Specifically, HRT first autoregressively generates a discontinuous sequence (e.g., make a prediction every $k$ tokens, $k>1$) and then fills in all previously skipped tokens at once in a non-autoregressive manner. Thus, we can easily trade off the translation quality and speed by adjusting $k$. In addition, by integrating other modeling techniques (e.g., sequence-level knowledge distillation and deep-encoder-shallow-decoder layer allocation strategy) and a mass of engineering efforts, HRT improves 80\% inference speed and achieves equivalent translation performance with the same-capacity AT counterpart.
Our fastest system reaches 6k+ words/second on the GPU latency setting, estimated to be about 3.1x faster than the last year's winner. 
\end{abstract}

\section{Introduction}

Large-scale transformer models have made impressive progress in past WMT translation tasks, but it is still challenging for practical model deployment due to time-consuming inference speed \cite{wang-etal-2018-niutrans,li-etal-2019-niutrans}. To build a fast and accurate machine translation system, participants in past WMT efficiency tasks developed and validated many efficient techniques, such as knowledge distillation \cite{hinton2015distilling, kim2016sequence}, light network architecture \cite{kasai2020deep}, quantization \cite{Lin2020TowardsF8} etc. We noticed that all the above efforts are aimed at autoregressive translation (AT) models. In contrast, other translation paradigms, like non-autoregressive translation (NAT) \cite{gu2017non} or semi-autoregressive translation (SAT) \cite{wang2018semi} etc., have not been well studied.

\newcommand{\masktext}[1]{\color{bluemunsell}{\textbf{#1}\xspace}}
\newcommand{\step}{\color{red}{$\rightarrow$\xspace}}
\begin{table*}[t]
	{
		\centering
		\resizebox{\textwidth}{!}
		{
			\begin{tabular}{L{.16\linewidth} M{.85\linewidth}}
				\toprule[1pt]
				\textbf{Source} & 
				\_\_The \_\_Next \_\_Big \_\_Labor \_\_ Strike \_\_Hit s \_\_Oregon \\
				
				\midrule
				\textbf{AT} & \_\_Der {\step} \_\_nächste {\step} \_\_große {\step} \_\_Arbeits {\step} streik {\step} \_\_trifft {\step} \_\_Oregon {\step} \eos \\
				\textbf{SAT} & \_\_Der \_\_nächste {\step}  \_\_große \_\_Arbeits {\step} streik \_\_trifft {\step} \_\_Oregon {\eos} \\
				\textbf{NAT} & {\masktext{\_\_Der}} {\masktext{\_\_nächste}}  {\masktext{\_\_große}} {\masktext{\_\_Arbeits}} {\masktext{streik}} {\masktext{\_\_trifft}} {\masktext{\_\_Oregon}} {\masktext{\eos}} \\
				
				\textbf{HRT (Stage \uppercase\expandafter{\romannumeral1})} & \_\_nächste {\step} \_\_Arbeits {\step} \_\_trifft {\step} \eos \\
				
				\textbf{HRT (Stage \uppercase\expandafter{\romannumeral2})}	& 
				{\masktext{\_\_Der}} \_\_nächste {\masktext{\_\_große}} \_\_Arbeits {\masktext{streik}} \_\_trifft {\masktext{\_\_Oregon}}  {\eos}    \\ 
				\bottomrule[1pt]
			\end{tabular}
		}
		\vspace{-.5em}
		\captionof{table}{Illustrations of different translation paradigms. \_\_ is the special symbol for whitespace in sentencepiece. {\step} denotes an autoregressive decoding step. {\masktext{Blue}} denotes that the token is generated in non-autoregressive way. }
		\label{table:case_study}	
		\vspace{-.5em}
	}
\end{table*}

In this participation, we restrict ourselves to the GPU latency track and attempt to investigate the potential of non-standard translation paradigms.
However, replicating the vanilla non-autoregressive or semi-autoregressive models degrades the translation quality severely in our preliminary experiments. To this end, we explore hybrid-regressive translation (HRT), the two-stage translation prototype, to better combine the advantages of autoregressive and non-autoregressive translation \cite{wang2021hybridregressive}. Specifically, HRT first uses an autoregressive decoder to generate a discontinuous target sequence with the interval $k$ ($k>1$). Then, HRT fills the co at once with a non-autoregressive decoder. The two decoders share the same parameters without adding additional ones. Thus, HRT can easily trade-off between translation quality and speed by adjusting $k$ \footnote{A larger $k$ implies that fewer autoregressive decoding steps are required, resulting in faster inference speed but lower translation quality.}. Please see Table~\ref{table:case_study} for the comparison between different translation paradigms.

In addition to the change of translation paradigm, we have also made a mass of other optimizations. We use the widely used sequence-level knowledge distillation \cite{kim2016sequence} and deep-encoder-shallow-decoder layer allocation strategy \cite{kasai2020deep} to learn effective compact models. Moreover, on the engineering side, we customized an efficient implementation of GPU memory reuse and kernel fusion for HRT following LightSeq  \cite{wang-etal-2021-lightseq}.

\begin{figure*}[t]
	\begin{center}
		\setlength{\tabcolsep}{2pt}

		\begin{tabular}{C{.24\textwidth}C{.24\textwidth}C{.24\textwidth}C{.24\textwidth}}
			
		\subfloat [\small{\texttt{AT}}] 
		{
		\begin{tikzpicture}
		\begin{scope}
		\setlength{\vseg}{2.5em}
		\setlength{\hseg}{0.6em}
		\setlength{\wnode}{2.5em}
		\setlength{\hnode}{1.5em}
		\tikzstyle{inputnode} = [rectangle, draw, thin, rounded corners=2pt, inner sep=1pt, fill=red!20, minimum height=0.7\hnode, minimum width=0.5\wnode]
		\tikzstyle{decnode} = [fill=blue!20,minimum width=3.8*\wnode, draw=blue!50,rounded corners=2pt]
		\tikzstyle{outputnode} = [rectangle, draw, thin, rounded corners=2pt, inner sep=1pt, fill=ugreen!20, minimum height=0.7\hnode, minimum width=0.5\wnode]
		
		\node [inputnode, anchor=west] (i1) at (0,0) {\footnotesize{\shortbos} };
		\node [outputnode, anchor=south] (o1) at ([yshift=\vseg] i1.north) {\footnotesize{$y_1$} };
		\draw [->,thin] (i1.north) -- ([yshift=-1pt] o1.south);
		
		\foreach \p/\i/\x/\y in {1/2/$y_1$/$y_2$,2/3/$y_2$/$y_3$,3/4/$y_3$/$y_4$,4/5/$y_4$/\shorteos}
		{
			\node [inputnode, anchor=west] (i\i) at ([xshift=\hseg] i\p.east) {\footnotesize{\x} };	
			\node [outputnode, anchor=south] (o\i) at ([yshift=\vseg] i\i.north) {\footnotesize{\y} };
		    \draw [->,thin] (i\i.north) -- ([yshift=-1pt] o\i.south);
		}
		\node[decnode] at ([yshift=0.5\vseg] i3.north) {\footnotesize{Skip-AT Decoder}};
		\end{scope}			
		\end{tikzpicture}	
		} &
	    \subfloat [\small{\texttt{CMLM}}]
	    {
		\begin{tikzpicture}
		\begin{scope}
		\setlength{\vseg}{2.5em}
		\setlength{\hseg}{.5em}
		\setlength{\wnode}{2.5em}
		\setlength{\hnode}{1.5em}
		\tikzstyle{inputnode} = [rectangle, draw, thin, rounded corners=2pt, inner sep=0pt, fill=red!20, minimum height=0.7\hnode, minimum width=0.5\wnode]
		\tikzstyle{decnode} = [fill=mypurple!20,minimum width=3.8*\wnode, draw=mypurple!50,rounded corners=2pt]
		\tikzstyle{outputnode} = [rectangle, draw, thin, rounded corners=2pt, inner sep=0pt, fill=ugreen!30, minimum height=0.7\hnode, minimum width=0.5\wnode]
		
		\node [inputnode, anchor=west] (i1) at (0,0) {\footnotesize{$y_1$} };
		\node [outputnode, anchor=south] (o1) at ([yshift=\vseg] i1.north) {\footnotesize{\spe{P}} };
		\draw [->,thin] (i1.north) -- ([yshift=-1pt] o1.south);
		
		\foreach \p/\i/\x/\y in {1/2/\shortmask/$y_2$,2/3/\shortmask/$y_3$,3/4/$y_4$/\shortpad,4/5/\shorteos/\shortpad}
		{
			\node [inputnode, anchor=west] (i\i) at ([xshift=\hseg] i\p.east) {\footnotesize{\x} };					
			\node [outputnode, anchor=south] (o\i) at ([yshift=\vseg] i\i.north) {\footnotesize{\y} };
		
		    \draw [->,thin] (i\i.north) -- ([yshift=-1pt] o\i.south);
		}
		\node[decnode] at ([yshift=0.5\vseg] i3.north) {\footnotesize{Skip-CMLM Decoder}};
		
		\end{scope}			
		\end{tikzpicture}

		} &
		\subfloat [\small{\texttt{SKIP-AT}}]
		{
		\begin{tikzpicture}
		\begin{scope}
		\setlength{\vseg}{2.5em}
		\setlength{\hseg}{.9em}
		\setlength{\wnode}{2.5em}
		\setlength{\hnode}{1.5em}
		\tikzstyle{inputnode} = [rectangle, draw, thin, rounded corners=2pt, inner sep=1pt, fill=red!20, minimum height=0.7\hnode, minimum width=0.5\wnode]
		\tikzstyle{decnode} = [fill=blue!20,minimum width=3.7*\wnode, draw=blue!50,rounded corners=2pt]
		\tikzstyle{outputnode} = [rectangle, draw, thin, rounded corners=2pt, inner sep=1pt, fill=ugreen!30, minimum height=0.7\hnode, minimum width=0.5\wnode]
		
		\node [inputnode, anchor=west] (i1) at (0,0) {\footnotesize{\spe{B$_2$}} };
		\node [outputnode, anchor=south] (o1) at ([yshift=\vseg] i1.north) {\footnotesize{$y_2$} };
		\draw [->,thin] (i1.north) -- ([yshift=-1pt] o1.south);
		
		\foreach \p/\i/\x/\y in {1/2/$y_2$/$y_4$,2/3/$y_4$/\shorteos}
		{
			\node [inputnode, anchor=west] (i\i) at ([xshift=2\hseg] i\p.east) {\footnotesize{\x} };					
			\node [outputnode, anchor=south] (o\i) at ([yshift=\vseg] i\i.north) {\footnotesize{\y} };
		
		    \draw [->,thin] (i\i.north) -- ([yshift=-1pt] o\i.south);
		}
		\node[decnode] at ([yshift=0.5\vseg] i2.north) {\footnotesize{Skip-AT Decoder}};
		
		\end{scope}			
		\end{tikzpicture}

		} &
			
		\subfloat [\small{\texttt{SKIP-CMLM}}]
		{
		\begin{tikzpicture}
		\begin{scope}
		\setlength{\vseg}{2.5em}
		\setlength{\hseg}{.7em}
		\setlength{\wnode}{2.5em}
		\setlength{\hnode}{1.5em}
		\tikzstyle{inputnode} = [rectangle, draw, thin, rounded corners=2pt, inner sep=1pt, fill=red!20, minimum height=0.7\hnode, minimum width=0.5\wnode]
		\tikzstyle{decnode} = [fill=mypurple!20,minimum width=4.1*\wnode, draw=mypurple!50,rounded corners=2pt]
		\tikzstyle{outputnode} = [rectangle, draw, thin, rounded corners=2pt, inner sep=1pt, fill=ugreen!30, minimum height=0.7\hnode, minimum width=0.5\wnode]
		
		\node [inputnode, anchor=west] (i1) at (0,0) {\footnotesize{\spe{M}} };
		\node [outputnode, anchor=south] (o1) at ([yshift=\vseg] i1.north) {\footnotesize{$y_1$} };
		\draw [->,thin] (i1.north) -- ([yshift=-1pt] o1.south);
		
		\foreach \p/\i/\x/\y in {1/2/$y_2$/\shortpad,2/3/\shortmask/$y_3$,3/4/$y_4$/\shortpad,4/5/\shorteos/\shortpad}
		{
			\node [inputnode, anchor=west] (i\i) at ([xshift=\hseg] i\p.east) {\footnotesize{\x} };					
			\node [outputnode, anchor=south] (o\i) at ([yshift=\vseg] i\i.north) {\footnotesize{\y} };
		
		    \draw [->,thin] (i\i.north) -- ([yshift=-1pt] o\i.south);
		}
		\node[decnode] at ([yshift=0.5\vseg] i3.north) {\footnotesize{Skip-CMLM Decoder}};
		
		\end{scope}			
		\end{tikzpicture}
		}
		\\
			
		\end{tabular}
	\end{center}
	
	\begin{center}
		\vspace{-.5em}
		\caption{Examples of training samples for four tasks, in which (a) and (b) are auxiliary tasks and (c) and (d) are primary tasks. For the sake of clarity, we omit the source sequence. \shortbos/\shorteos/\shortpad/\shortmask represents the special token for \bos/\eos/\pad/\mask, respectively. \spe{B$_2$} is the \bos for k=2. Loss at \spe{P} is ignored. }
	\label{fig:train_sample}
		\vspace{-1.2em}
	\end{center}
\end{figure*}
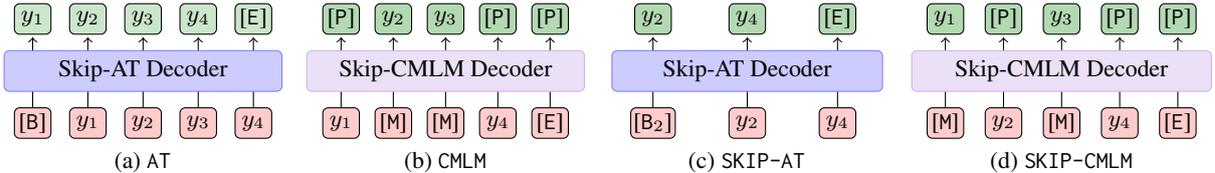

Putting all the efforts together, our HRT model achieves almost equivalent BLEU scores to the corresponding AT counterparts while improving the inference speed by about 80\%. Our best-BLEU model drops an average of 0.9 BLEU points compared to the teacher model, which ensembles four transformer-big models. Moreover, our fastest model decodes 6k+ source words per second, estimated to be 3.1x faster than the winner in last year \footnote{We obtained the best decoding speed in last year's competition according to \citet{heafield-etal-2021-findings}: The fastest system \textit{2.12\_1.micro.rowcol-0.5} decodes 19,951,184
space-separated words in 13665 seconds.
Therefore, we estimated its inference speed is 1460 words/second. We note that the acceleration ratio is not an accurate value because we use slightly different computation devices and test data to measure the speed.}.

\section{Hybrid-regressive translation}
One of the most important highlights is the introduction of the newly proposed two-stage translation prototype——HRT. In this section, we will detail the model, training, and decoding of HRT.

\subsection{Model}
HRT consists of three components: encoder, Skip-AT decoder (for stage \uppercase\expandafter{\romannumeral1}), and Skip-CMLM decoder (for stage \uppercase\expandafter{\romannumeral2}). All components adopt the Transformer architecture \citep{vaswani2017attention}. 
The two decoders have the same network structure, and we share them to make the parameter size of HRT the same as the vanilla Transformer.
The only difference between the two decoders lies in the masking pattern in self-attention: The Skip-AT decoder masks future tokens to guarantee strict left-to-right generation like an autoregressive Transformer \cite{vaswani2017attention}. In contrast, the Skip-CMLM decoder eliminates it to leverage the bi-directional context like the standard conditional masked language model (CMLM) \cite{ghazvininejad-etal-2019-mask}. We note that there is no specific target length prediction module in HRT because HRT can obtain the translation length as the by-product of the Skip-AT decoder: $N_{nat}$=$k \times N_{at}$, where $N_{at}$ is the sequence length produced by Skip-AT.

\subsection{Training}

\paragraph{Multi-task framework.}
We learn HRT through joint training of four tasks, including two primary tasks (\texttt{SKIP-AT}, \texttt{SKIP-CMLM}) and two auxiliary tasks (\texttt{AT}, \texttt{CMLM}).
All tasks use cross-entropy as the training objective.
Figure~\ref{fig:train_sample} illustrates the differences in training samples among these tasks.
It should be noted that, compared with \texttt{AT}, \texttt{SKIP-AT} shrinks the sequence length from $N$ to $N/k$, whereas the token positions follow the original sequence. For example, in Figure~\ref{fig:train_sample} (c), the position of Skip-AT input (\spe{B$_2$}, $y_2$, $y_4$) is (0, 2, 4) instead of (0, 1, 2).
Involving auxiliary tasks is necessary because the two primary tasks cannot fully leverage all tokens in the sequence due to the fixed $k$.
For example, in Figure~\ref{fig:train_sample} (c) and (d), $y_1$ and $y_3$ have no chance to be learned as the decoder input of either \texttt{SKIP-AT} or \texttt{SKIP-CMLM}.

\paragraph{Curriculum learning.}
To ensure that the model is not overly biased towards auxiliary tasks, we propose gradually transferring the training tasks from auxiliary tasks to primary tasks through curriculum learning \citep{bengio-cl}.  
More concretely, given a batch of original sentence pairs $\mathcal{B}$, and the proportion of primary tasks in $\mathcal{B}$ is $p_k$, we start with $p_k$=0 and construct the training samples of \texttt{AT} and \texttt{CMLM} for all pairs. 
Then we gradually increase $p_k$ to introduce more learning signals for \texttt{SKIP-AT} and \texttt{SKIP-CMLM} until $p_k$=1. In implementation, we schedule $p_k$ by:
\begin{equation}
	\label{eq:skip_rate}
	p_k = ( t/T )^\lambda,
\end{equation}
where $t$ and $T$ are the current and total training steps. $\lambda$ is a hyperparameter, and we use $\lambda$=1 to increase $p_k$ linearly for all experiments.

\subsection{Decoding}

HRT adopts two-stage generation strategy: In the first stage, the Skip-AT decoder starts from \spe{BOS$_k$} to autoregressively generate a discontinuous target sequence $\hat{\vy}_{at}=(z_1, z_2, \ldots, z_m)$ with chunk size $k$ until meeting \eos. Then we construct the input of Skip-CMLM decoder $\vy_{nat}$ by appending $k-1$ [\texttt{MASK}]s before every $z_i$. The final translation is generated by replacing all [\texttt{MASK}]s with the predicted tokens by the Skip-CMLM decoder with one iteration. If there are multiple [\texttt{EOS}]s existing, we truncate to the first \eos. 
Note that the beam size b$_{at}$ in Skip-AT can be different from the beam size b$_{nat}$ in Skip-CMLM as long as st. b$_{at}$ $\ge$ b$_{nat}$: We only feed the top b$_{nat}$ Skip-AT hypothesis to Skip-CMLM decoder. 
Finally, we choose the translation hypothesis with the highest score $\rm{S}(\hat{\vy})$ by:
\begin{equation}
\label{eq:inf_score}
\resizebox{.45\textwidth}{!}
{$
\begin{aligned}
 \underbrace{\sum_{i=1}^{m}{\log{P}(z_i|\vx,\vz_{<i})}}_{\text{Skip-AT score}} +	\underbrace{\sum_{i=0}^{m-1}\sum_{j=1}^{k-1}{\log{P}(\hat{y}_{i \times k+j}|\vx,\vy_{nat})}}_{\text{Skip-CMLM score}}
\end{aligned}
$}    
\end{equation}
where $z_i$=$\hat{y}_{i \times k}$.

\section{Optimization}

\paragraph{Sequence-level knowledge distillation.} 
Overall, we use the teacher-student framework via sequence-level knowledge distillation (\textsc{seqkd}) to learn our small HRT model \cite{kim2016sequence}.
Specifically, the ensemble of provided four transformer-big models is our teacher, whose beam search results are used as our distillation data. There are 320M official distillation data composed of 80M parallel and 240M monolingual datasets. We directly use the distillation data without further data cleaning. We use the same sentencepiece vocabulary as the teacher model to encode the text.

\paragraph{Deep-encoder-shallow-decoder architecture.}

\begin{table}[t]
	{
		\centering
			\begin{tabular}{c c c c}
				\toprule[1pt]
				\textbf{Encoder} & \textbf{Newstest19}  &  \textbf{Newstest20} & \textbf{WPS} \\ \midrule
				6 & 43.8 & 32.6 & 4.0k  \\
				12 & 45.6 & 33.9 & 3.8k  \\
				20 & 45.9 & 34.4 & 3.5k  \\
				
				\bottomrule[1pt]
			\end{tabular}
		\vspace{-.5em}
		\captionof{table}{SacreBLEU and inference speed against the number of encoder layers in HRT with a single-layer decoder. All HRT models are trained with $k$=2. WPS refers to source words per second, measured by the average five runs with a batch size of 1. Unless otherwise stated, we measure WPS on Newstest20.}
		\label{table:encoder_layer}	
		\vspace{-.5em}
	}
\end{table}

Using deep-encoder-shallow-decoder network architecture has been widely validated effectiveness for transformer-based NMT systems \cite{wang-etal-2021-niutrans,kasai2020deep}. Our HRT also follows this guidance by using only one decoder layer. We use the pre-norm transformer following \citet{wang2019learning} to learn deep encoder well. Intuitively, the single-layer decoder may be insufficient for HRT because the decoder is responsible for both autoregressive and non-autoregressive generation. However, as shown in Table~\ref{table:encoder_layer}, we found that HRT enjoys the deep-encoder-shallow-decoder architecture. 
For example, compared to HRT\_E6D1 \footnote{We use HRT\_E\{\#1\}D\{\#2\} to denote the HRT model with \{\#1\}-layer encoder and \{\#2\}-layer decoder.}, HRT\_E12D1 and HRT\_E20D1 improve +1.6/+2.0 BLEU score points on average, while the inference speed decreases by 5\% and 12.5\%. Therefore, we mainly investigate HRT with a 12-layer and 20-layer encoder due to the high BLEU scores.

\paragraph{Fully greedy search.}

\begin{table}[t]
	{
		\centering
			\begin{tabular}{c c c c}
				\toprule[1pt]
				\textbf{b$_{at}$} & \textbf{b$_{nat}$}  &  \textbf{Newstest19} & \textbf{Speedup} \\ \midrule
				5 & 5 & \textbf{45.9} & ref.  \\
				5 & 1 & 45.8 & 1.05x  \\
				1 & 1 & 45.6 & \textbf{1.33x}  \\
				
				\bottomrule[1pt]
			\end{tabular}
		\vspace{-.5em}
		\captionof{table}{Effects of different settings of beam size in HRT. }
		\label{table:beam_size}	
		\vspace{-.5em}
	}
\end{table}

\begin{table*}[t]
	\begin{center}
	\resizebox{0.95\textwidth}{!}
	{
	\begin{tabular}{l c c c c c c c c c c c}
		\toprule[1pt]
        
        \multicolumn{1}{c}{\multirow{2}{*}{\textbf{Model}}} & 
        \multicolumn{1}{c}{\multirow{2}{*}{\textbf{\#Param.}}} & 
        \multicolumn{2}{c }{\textbf{Newstest19}} & 	\phantom{a} & 
		\multicolumn{2}{c }{\textbf{Newstest20}} &\phantom{a} &
		\multicolumn{2}{c}{\textbf{Average}} & \phantom{a} & 
		\multicolumn{1}{c}{\multirow{2}{*}{\textbf{WPS}}} \\
		\cmidrule{3-4} \cmidrule{6-7} \cmidrule{9-10}
		
		~ & ~ & \textbf{BLEU} & \textbf{COMET} & & \textbf{BLEU} & \textbf{COMET} & & \textbf{BLEU} & \textbf{COMET} && ~ \\
		\midrule
		
		Teacher (four transformer-big) & 4$\times$209.1M & 47.1 & - && 35.0 & - && 41.1 & - && - \\
		WMT21 fastest \cite{behnke-etal-2021-efficient} & 9.0M & - & - && 33.3 & - && - & - && 1.5k$^*$ \\
		\midrule
		AT\_E6D1 & 39.5M & 45.1 & 0.551 && 33.9 & 0.469 && 39.5 & 0.510 && 2.2k \\
		AT\_E12D1 & 58.4M & 45.4 & 0.572 && 34.1 & 0.489 && 39.8 & 0.530 && 2.1k \\
		AT\_E20D1 & 83.6M & 45.9 & 0.581 && 34.6 & 0.502 && 40.3 & 0.541 && 1.9k \\
		\midrule
		HRT\_E12D1 (k=2) & 58.4M & 45.6 & 0.547 && 33.9 & 0.454 && 39.8 & 0.500 && 3.8k \\
		HRT\_E12D1 (k=3) & 58.4M & 45.0 & 0.503 && 33.6 & 0.377 && 39.3 & 0.440 && 4.9k \\
		HRT\_E12D1 (k=4) & 58.4M & 44.1 & 0.432 && 32.9 & 0.267 && 38.5 & 0.350 && \textbf{6.1k}  \\
		\midrule
		HRT\_E20D1 (k=2) & 83.6M & 45.9 & 0.561 && 34.4 & 0.472 && \textbf{40.2} & \textbf{0.517} && 3.5k \\
		HRT\_E20D1 (k=3) & 83.6M & 45.3 & 0.524 && 34.0 & 0.406 && 39.7 & 0.465 && 4.5k \\
		HRT\_E20D1 (k=4) & 83.6M & 44.2 & 0.435 && 33.3 & 0.283 && 38.8 & 0.360 && 5.4k \\
		
	\bottomrule[1pt]
	\end{tabular}
	}
		
	\caption{Compare different model variants with regard to SacreBLEU \cite{post-2018-call}, COMET \cite{rei-etal-2020-comet} and inference speed. $*$ denotes the number is not exactly comparable due to the difference in test data and GPU.}
	\label{table:main_results}
	\end{center}
\end{table*}

Prior work has validated that greedy search is sufficient for the autoregressive distilled model to work well \cite{kim2016sequence}. Since HRT refers to two beam sizes (b$_{at}$ and b$_{nat}$), we test three settings as shown in Table~\ref{table:beam_size}. It can be seen that using b$_{at}$=1 and b$_{nat}$=1 only decreases BLEU slightly but accelerates a 30\%+ faster than that of b$_{at}$=5 and b$_{nat}$=5. Unless otherwise stated, we use b$_{at}$=1 and b$_{nat}$=1 in the following experiments.

\paragraph{Maximum sequence length.}
We predefine the maximum source/target sequence length $L$ as 200. Here the length is calculated based on the results of sentencepiece. Once the sequence length exceeds $L$, we truncate the source/target sequence.
For HRT, we let the maximum decoding length in the Skip-AT stage as $L/k$. In this way, the maximum target length in the Skip-CMLM stage can be guaranteed not beyond $L$. 

\paragraph{GPU memory reuse.}
Since we only participate in the GPU latency track, given the predefined maximum sequence length $L$, we can estimate the maximum GPU memory buffer used in the encoder, autoregressive decoder, and non-autoregressive decoder in advance, respectively. Then we only allocate the maximum buffer size among them because these three processes are memory-independent. This memory-reuse method helps us reduce our footprints and avoid frequent memory applications and releases.

\paragraph{Kernel fusion.}
Too many fine-grained kernel functions make modern GPU inefficient due to kernel launching overhead and frequent memory I/O addressing \cite{wang-etal-2021-lightseq,wu-etal-2021-tentrans}. 
We follow the good implementation in LightSeq and use the general matrix multiply (GEMM) provided by cuBLAS as much as possible, with some custom kernel functions. Please refer to \citet{wang-etal-2021-lightseq} for details.

\paragraph{FP16 inference.}

\begin{table}[t]
	{
		\centering
			\begin{tabular}{c c c}
				\toprule[1pt]
				\textbf{Model} & \textbf{FP16} & \textbf{WPS} \\ \midrule
				HRT\_E12D1 (k=2) & no & 3.3k \\
				HRT\_E12D1 (k=2) & yes & 3.8k \\
				
				\bottomrule[1pt]
			\end{tabular}
		\vspace{-.5em}
		\captionof{table}{The effect of FP16 on HRT model in GPU latency task.}
		\label{table:fp16}	
		\vspace{-.5em}
	}
\end{table}

We also use the 16-bit floating-point to utilize modern GPU hardware efficiently. Previous study \cite{wang-etal-2021-niutrans} shows that FP16 can bring significant acceleration in batch decoding. In contrast, in our GPU latency task, we only observed about 15\% speedup due to the smaller computational burden, as shown in Table~\ref{table:fp16}.

\paragraph{Docker submission.}
We use multistage builds to reduce the docker image size. Specifically, we first use static compilation to build our executable program with CUDA 11.2. Then we add the built result and model into the 11.2.0-base-centos7 docker. The model disk size is compressed by \textit{xz} compression toolkit.

\section{Experimental Results}

\paragraph{Setup.}
We mainly compared HRT to the standard autoregressive baselines in Table~\ref{table:main_results}. All models adopt transformer-base setting \cite{vaswani2017attention}: $d$=512, $d_{ff}$=2048, $head$=8. We validated the following model variants:
\begin{itemize}
	\item \textbf{AT:} We train three autoregressive baselines with the number of encoder layers of 6, 12, and 20, denoted as AT\_E6D1, AT\_E12D1, and AT\_E20D1, respectively. All AT models are trained from scratch for 300k steps.  
	\item \textbf{HRT:} HRT models are fine-tuned based on the pre-trained AT counterparts for 300k steps. We also try different chunk sizes $k \in \{2, 3, 4\}$ to trade off the translation quality and inference speed.  
\end{itemize}
Other training hyper-parameters are the same as \citet{wang2019learning}. We ran all experiments on 8 GeForce 3090 GPUs. For decoding, the length penalty is 0.6, and the batch size is 1. 
We report the detokenized SacreBLEU score with the same signature as the teacher. Besides, we also follow \citet{helcl-etal-2022-non}'s suggestion to provide COMET score \cite{rei-etal-2020-comet} for the evaluation of non-autoregressive translation.

\paragraph{Translation quality.} First, we can see that a deeper encoder improves about 0.5 BLEU points across the board. 
When using $k$=2 for HRT, both 12-layer and 20-layer HRT models have almost equivalent BLEU scores to that of AT counterparts. Our best HRT model HRT\_E20D1 with $k$=2 only drops an average of 0.9 BLEU points than the teacher using model ensemble.
However, in line with \citet{helcl-etal-2022-non}, we find that even when BLEU scores are close, HRT's COMET scores are significantly lower than those of AT, e.g., AT\_E20D1 vs. HRT\_E20D1 (k=2). Nevertheless, HRT\_E20D1 (k=2) still achieves higher BLEU and COMET than AT\_E6D1 with 60\% acceleration.

\paragraph{Translation speed.} We estimated the inference speed of the fastest system last year according to the data in \citet{heafield-etal-2021-findings}. Supposing ignoring the difference in test data, our AT baselines run about 40\%+ faster than it. It indicates that our AT engine is a strong baseline. Even so, we can see that both 12-layer and 20-layer HRT with k=2 achieve approximated 80\% acceleration than AT without BLEU drop. Moreover, larger $k$ further reduces the autoregressive decoding steps: Our fastest model, HRT\_E12D1 ($k$=4), decodes 6k+ source words/second, which is 3.1 times faster than the fastest system last year.

\section{Conclusion}

This paper presented the \myname system to the GPU latency track of the WMT 2022 translation efficiency task. We proposed hybrid-regressive translation, a novel two-stage prototype to replace conventional autoregressive translation. With a lot of development optimization, we showed that our HRT with a chunk size of 2 achieves equivalent translation performance to the AT counterpart while accelerating 80\% inference speed. By increasing HRT's chunk size, our system can further speed up 60\% to 6k+ words/second, estimated to be about 3.1 times faster than the fastest system in last year's competition.

\bibliography{anthology,custom}
\bibliographystyle{acl_natbib}


\end{document}